\documentclass[smallextended]{svjour3}       
\smartqed  
\usepackage{appendix}
\usepackage{amsmath}
\usepackage{amssymb}
\usepackage{graphicx}
\usepackage{array}
\usepackage{longtable}
\usepackage{natbib}
\usepackage{hyperref}
\usepackage{framed,multirow}

\usepackage{epsfig}
\usepackage[export]{adjustbox}
\usepackage{array}
\usepackage{rotating}
\usepackage{marvosym}
\usepackage{booktabs}
\usepackage{xspace}
\usepackage{csquotes}

\usepackage{multirow}
\usepackage{color}
\usepackage{color, colortbl}
\usepackage{float}
\usepackage[table,xcdraw]{xcolor}

\newcommand{\latinphrase}[1]{\textit{#1}} 

\newcommand{\ie}{\latinphrase{i.e.}\xspace}

\newcommand{\eg}{\latinphrase{e.g.}\xspace}
\newcommand{\etc}{\latinphrase{etc.}\xspace}

\definecolor{Gray}{gray}{0.85}
\newcommand*\patchAmsMathEnvironmentForLineno[1]{%
\expandafter\let\csname old#1\expandafter\endcsname\csname #1\endcsname
\expandafter\let\csname oldend#1\expandafter\endcsname\csname end#1\endcsname
\renewenvironment{#1}%
{\linenomath\csname old#1\endcsname}%
{\csname oldend#1\endcsname\endlinenomath}}%
\newcommand*\patchBothAmsMathEnvironmentsForLineno[1]{%
\patchAmsMathEnvironmentForLineno{#1}%
\patchAmsMathEnvironmentForLineno{#1*}}%
\AtBeginDocument{%
\patchBothAmsMathEnvironmentsForLineno{equation}%
\patchBothAmsMathEnvironmentsForLineno{align}%
\patchBothAmsMathEnvironmentsForLineno{flalign}%
\patchBothAmsMathEnvironmentsForLineno{alignat}%
\patchBothAmsMathEnvironmentsForLineno{gather}%
\patchBothAmsMathEnvironmentsForLineno{multline}%
}

\begin{document}

\title{A Systematic Evaluation: Fine-Grained CNN vs. Traditional CNN Classifiers
}

\author{Saeed Anwar$^*$         \and
        Nick Barnes \and
        Lars Petersson
}

\institute{$^*$Corresponding Author: Saeed Anwar \at Commonwealth Scientific Industrial Research Organization, Australia,\\
School of Engineering and Computer Science, The Australian National University, Australia,\\ Faculty of Engineering and Information Technology, The University of Technology Sydney, Australia.\\    
\email{saeed.anwar@(csiro.au, anu.edu.au, uts.edu.au)}
\and
Nick Barnes \at School of Engineering and Computer Science, Australian National University, Australia\\
\email{nick.barnes@anu.edu.au}\and
Lars Petersson \at Data61-Commonwealth Scientific Industrial Research Organization, Australia.\\
\email{lars.petersson@data61.csiro.au}
}


\maketitle

\begin{abstract}
Fine-grained classifiers collect information about inter-class variations to best use the underlying minute and subtle differences.  The task is very challenging due to the small differences between the colors, viewpoints, and structure in the same class entities. The classification becomes difficult due to the similarities between the differences in viewpoint with other classes and differences with its own.  This work investigates the landmark general CNN classifiers' performance, which presented top-notch results on large-scale classification datasets and fine-grained datasets and compared it against state-of-the-art fine-grained classifiers. This paper poses two specific questions: (i) Do the general CNN classifiers achieve comparable results to fine-grained classifiers? (ii) Do general CNN classifiers require any specific information to improve upon the fine-grained ones? We train the general CNN classifiers throughout this work without introducing any aspect specific to fine-grained datasets. We show an extensive evaluation on six datasets to determine whether the fine-grained classifier can elevate the baseline in their experiments. Train models and codes are available at \href{https://github.com/saeed-anwar/FGSE}{https://github.com/saeed-anwar/FGSE}
\keywords{Fine-grain visual classification\and Traditional classification\and Systematic Evaluation\and Deep learning\and Review\and  Baselines.}
\end{abstract}

\section{Introduction}
\label{intro}
Fine-grain visual classification (FGVC) refers to the task of distinguishing the categories of the same class. Fine-grain classification is different from  traditional classification as the former models intra-class variance while the later is about the inter-class difference. Examples of naturally occurring fine-grain classes include: birds \citep{WahCUBDataset,van2015NaBirdDataset}, dogs~\citep{khosla2011DogDataset}, flowers~\citep{nilsback2008FlowerDataset}, vegetables~\citep{hou2017vegfru}, plants~\citep{wegner2016cataloging} \etc while human-made categories include aeroplanes~\citep{maji2013aircraftdataset}, cars~\citep{krause2013CarDataset}, food~\citep{chen2009pfid} \etc Fine-grain classification is helpful in numerous computer vision and image processing applications such as image captioning~\citep{aafaq2019video}, machine teaching~\citep{spivak2012outside}, and instance segmentation~\citep{liu2018segment}, \etc

Fine-grain visual classification is a challenging problem as there are minute and subtle differences within the species of the same classes \eg, a crow and a raven, compared to traditional classification, where the difference between the classes is quite visible \eg, a lion and an elephant. Fine-grained visual classification of species or objects of any category is a herculean task for human beings and usually requires extensive domain knowledge to identify the species or objects correctly.

\begin{figure}
\begin{center}
\begin{tabular}{c@{ }c@{ }c}
    \includegraphics[width=2.6cm,height=2.6cm]{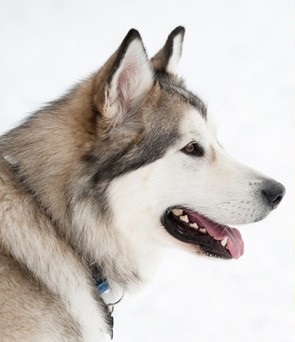}&  
    \includegraphics[width=2.6cm,height=2.6cm]{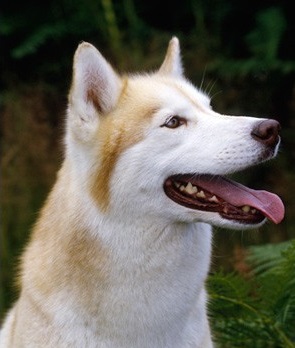}&  
    \includegraphics[width=2.6cm,height=2.6cm]{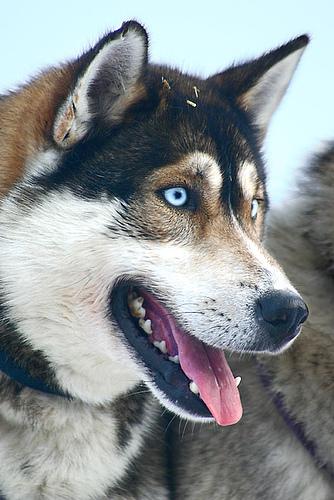}\\  
    Malamute & Husky& Eskimo\\
     \includegraphics[width=2.6cm,height=2.6cm]{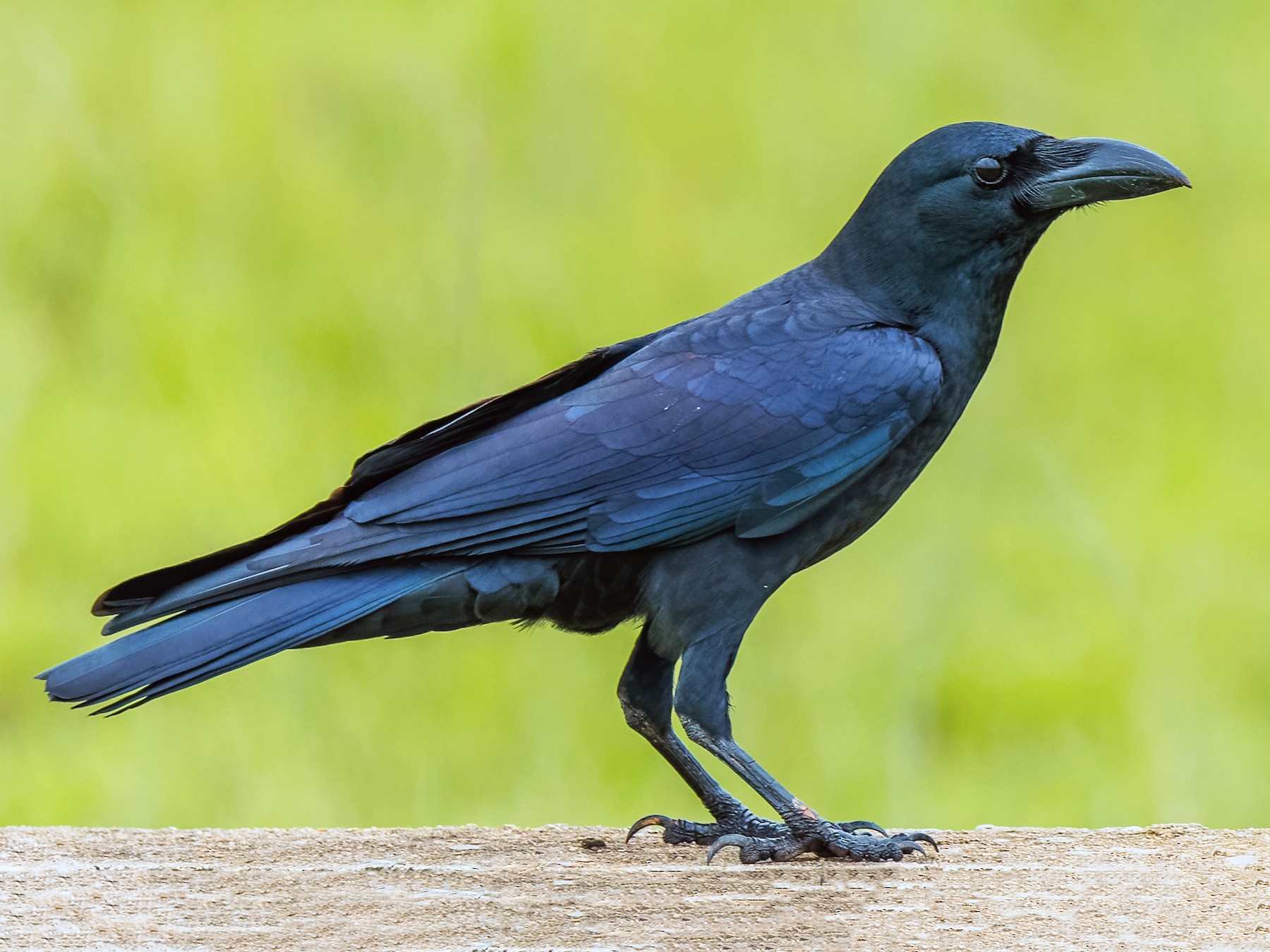}&  
     \includegraphics[width=2.6cm,height=2.6cm]{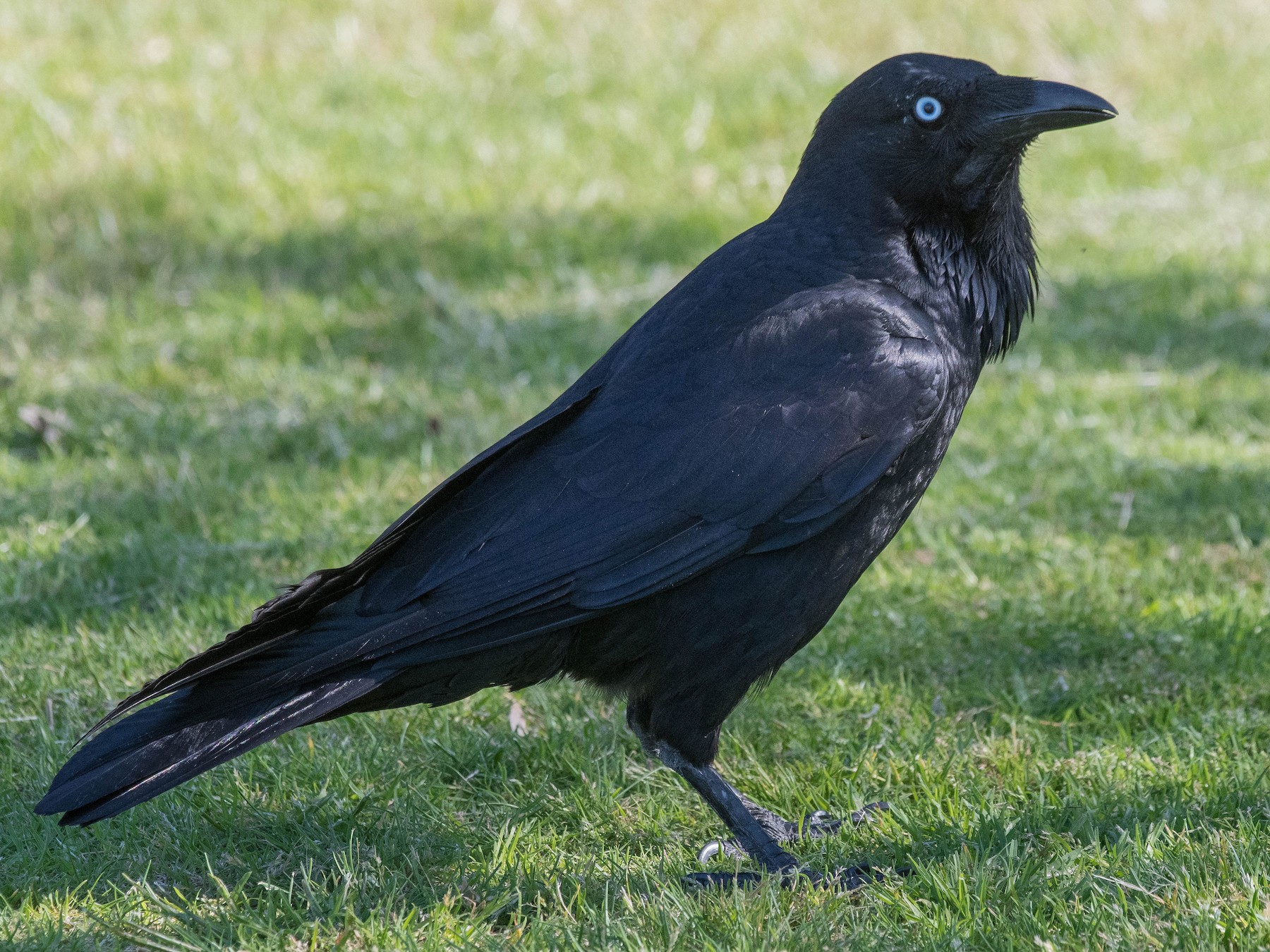}&  
     \includegraphics[width=2.6cm,height=2.6cm]{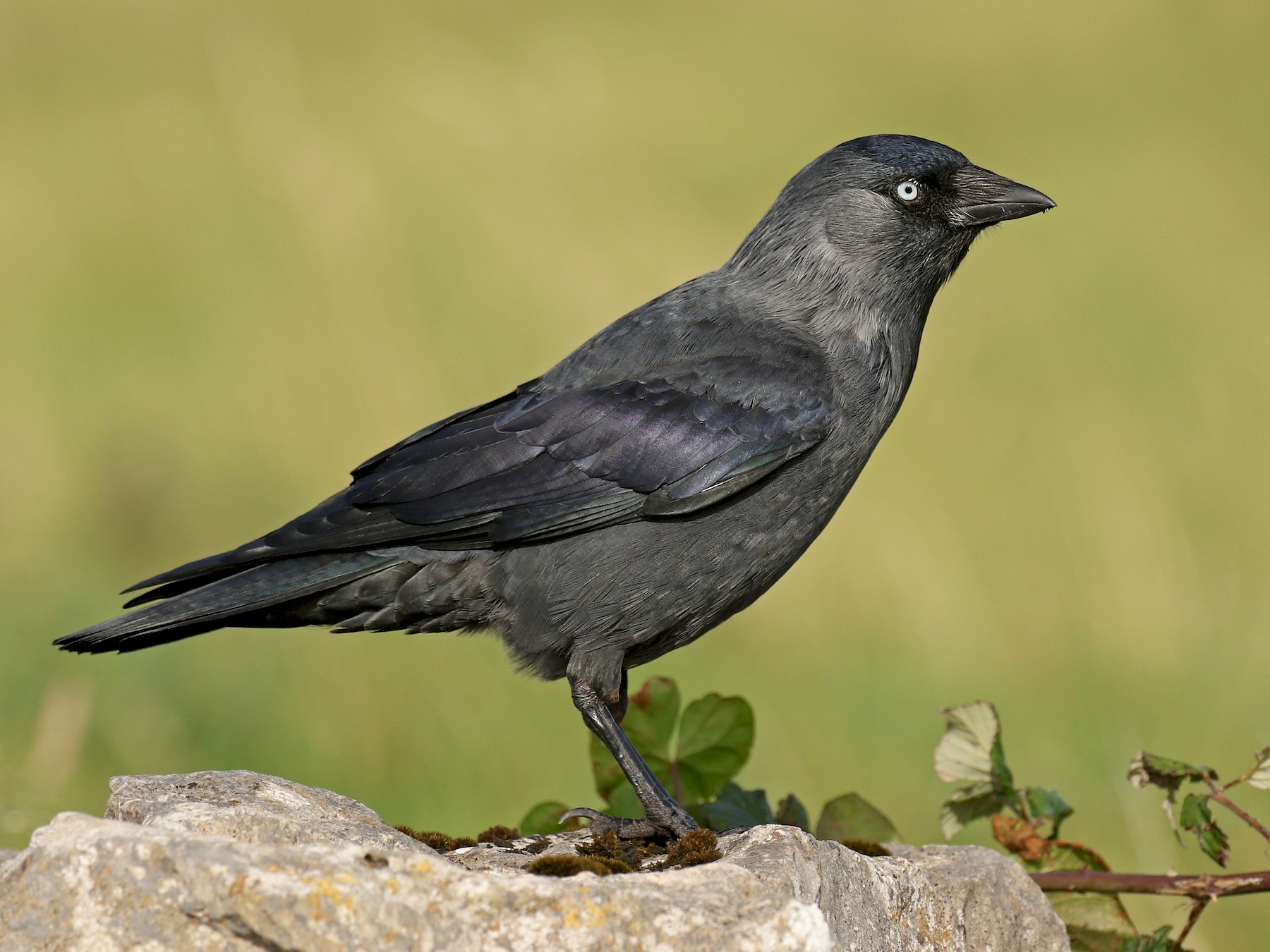}\\  
     Crow & Raven & Jackdaw\\
 \end{tabular}
\end{center}
\caption{The difference between classes (inter-class variation) is limited for various classes.}
\label{fig:cmp1}
\begin{center}
\begin{tabular}{c@{ }c@{ }c}
    \includegraphics[width=2.6cm,height=2.6cm]{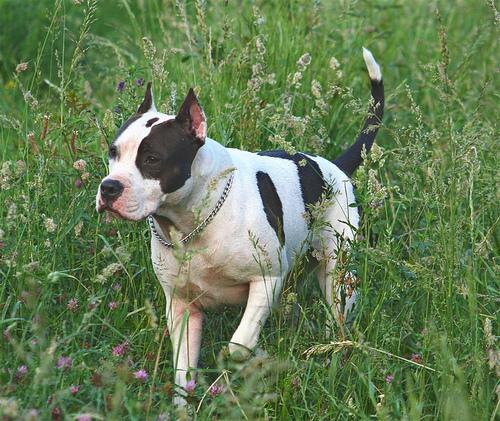}&  
    \includegraphics[width=2.6cm,height=2.6cm]{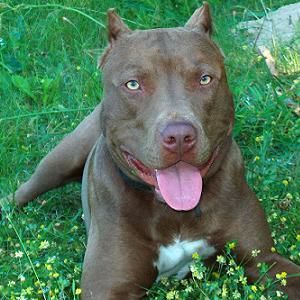}&  
    \includegraphics[width=2.6cm,height=2.6cm]{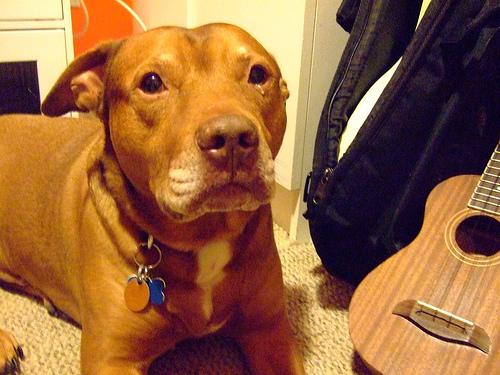}\\  
    
    \includegraphics[width=2.6cm,height=2.6cm]{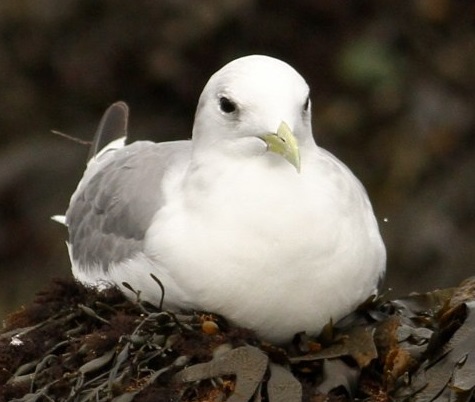}&  
    \includegraphics[width=2.6cm,height=2.6cm]{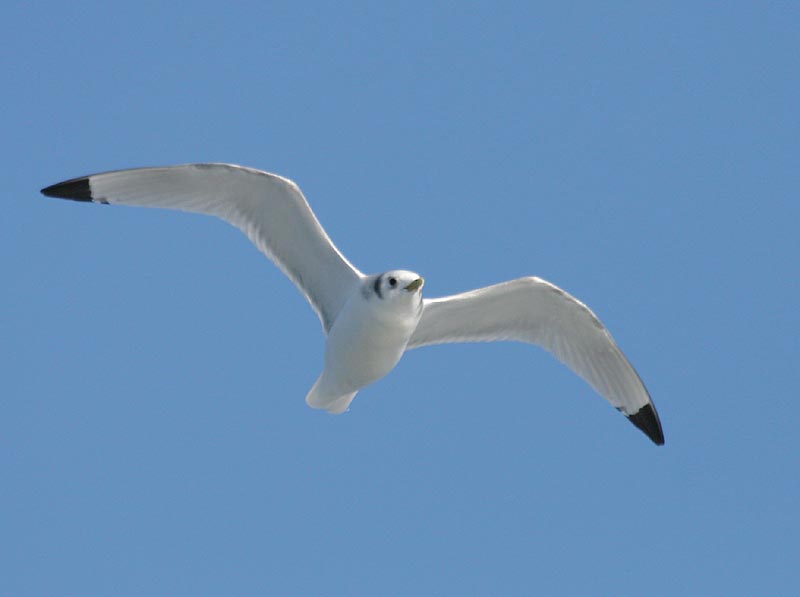}&  
    \includegraphics[width=2.6cm,height=2.6cm]{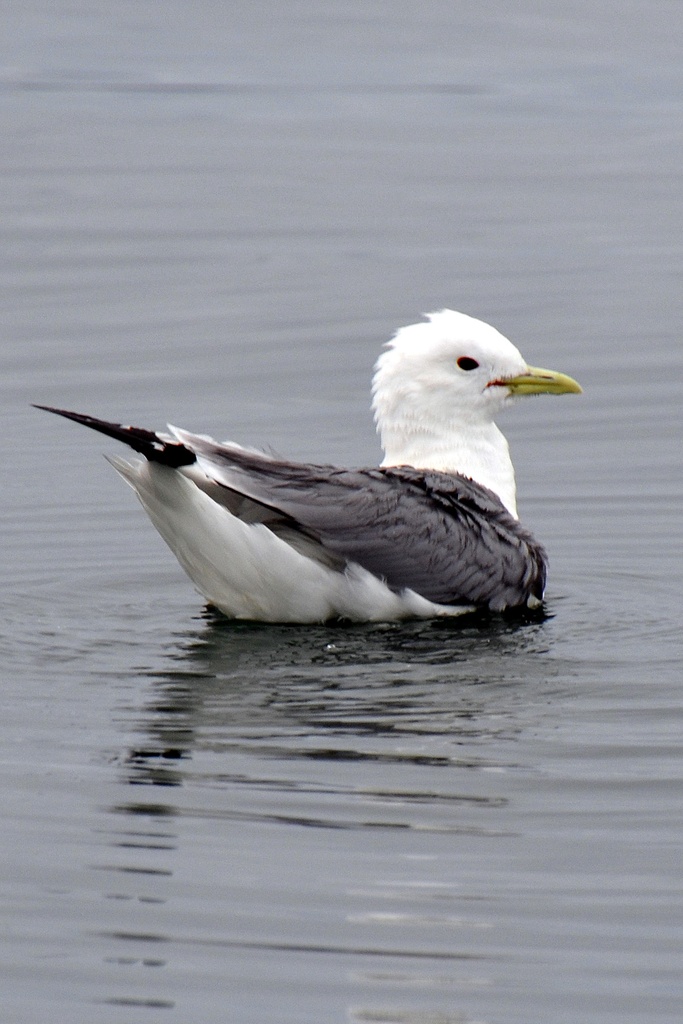}\\ 
 \end{tabular}
\end{center}
\caption{The intra-class variation is usually high due to pose, lighting, and color.}
\label{fig:cmp2}
\end{figure}

As mentioned earlier, fine-grained classification in image space aims to reduce the high intra-class variance and the low inter-class variance. We provide a few sample images from the dog and bird datasets in Figure~\ref{fig:cmp1} to highlight the difficulty of the problem. The examples in the figure show the images with the same viewpoint. The colors are also roughly similar. Although the visual variation is very limited between classes, all of these belong to different dog and bird categories. In Figure~\ref{fig:cmp2}, we provide more examples of the same mentioned categories.  Here, the difference in the viewpoint and colors are prominent. The visual variation is more significant than the images in Figure~\ref{fig:cmp1}, but these belong to the same class.

Many approaches have been proposed to tackle the problem of fine-grained classification; for example, earlier works converged on part detection to model the intra-class variations. Next, the algorithms exploited three-dimensional representations to hand multiple poses and viewpoints to achieve state-of-the-art results. Recently, with the advent of CNNs, most methods have exploited the modeling capacity of CNNs as a component or as a whole.
 
This paper aims to investigate the capability of traditional CNN networks compared to specially designed fine-grained CNN classifiers. We strive to answer whether current general CNN classifiers can achieve comparable performance to fine-grained ones. To show the competitiveness, we employ several fine-grained datasets and report top-1 accuracy for both classifier types. These experiments provide a proper place for general classifiers in fine-grained performance charts and serve as baselines for future comparisons for FGVC problems. 

 This paper is organized as: Section~\ref{sec:related} presents related work about the fine-grain classification networks. Section~\ref{sec:cnn} introduces the traditional state-of-the-art algorithms which will be compared against fine-grain classifiers. Section~\ref{sec:experiments} shows the experimental settings and datasets for evaluation. Section~\ref{sec:evaluations} offers a comparative evaluation between the traditional classifiers and fine-grain classifier; finally, section~\ref{sec:conclusion} concludes the paper.

\section{Fine-Grain Classifiers}
 \label{sec:related}
Fine-grained visual classification is an important and well-studied problem. Fine-grain visual classification aims to differentiate between subclasses of the same category instead of the traditional classification problem, where discriminative features are learned to distinguish between classes. Some of the challenges in this domain are the following: i) The categories are highly correlated \ie small differences and small inter-category variance to discriminate between subcategories.  ii): Similarly, the intra-category variation can be significant due to different viewpoints and poses. Many algorithms such as \citep{angelova2013image,chai2012tricos,deng2013fine,gavves2013fine,zhang2016picking,huang2016part,zhang2016embedding} are presented to achieve the desired results. In this section, we highlight the recent approaches. The FGVC research can be divided into the following main branches reviewed in the paragraphs below.

\textbf{Part-Based FGVC Algorithms}. The part-based category of algorithms relies on the distinguishing features of the objects to leverage the accuracy of visual recognition, which includes~\citep{yang2012unsupervised,chai2013symbiotic,berg2013poof,zhang2013deformable,zhang2014part,xiao2015application}. These FGVC methods~\citep{zheng2017learning,wang2015multiple} aim to learn the distinct features present in different parts of the object \eg the differences present in the beak and tail of the bird species. Similarly, the part-based approaches normalize the variation present due to poses and viewpoints. Many works~\citep{parkhi2011truth,parkhi2012cats,WahCUBDataset} assume the availability of bounding boxes at the object-level and the part-level in all the images during the training as well as testing settings. To achieve higher accuracy,~\citep{berg2013poof,liu2012dog,xie2013hierarchical} employed both object-level and part-level annotations. These assumptions restrict the applicability of the algorithms to larger datasets. A reasonable alternative setting would be the availability of a bounding box around the object of interest. Recently,~\cite{chai2013symbiotic} applied simultaneous segmentation and detection to enhance the performance of segmentation and object part localization. Similarly,  a supervised method is proposed \citep{gavves2013fine}, which locates the training images similar to a test image using KNN. The object part locations from the selected training images are regressed to the test image.

\textbf{Bounding Boxes Based Methods}. The succeeding supervised methods take advantage of the annotated data during the training phase while requiring no knowledge during the testing phase and learn on both object-level and object-parts level annotation in the training phase only.  Such an approach is furnished in \cite{krause2015fine}, where only object-level annotations are given during training while no supervision is provided at the object-parts level. Similarly, Spatial Transformer Network (STCNN)~\citep{jaderberg2015spatial} handles data representation and outputs vital regions' location. Furthermore, recent approaches focused on removing the limitation of previous works, aiming for conditions where the information about the object-parts location is not required either in the training or testing phase. These FGVC methods are suitable for deployment on a large scale and help the advancement of research in this direction. 

\textbf{Attention Models}. Recently, attention-based algorithms have been employed in FGVC, where the focus is on distinguishing parts via an attention mechanism. Using attention,~\cite{xiao2015application} presented two attention models to learn appropriate patches for a particular object and determine the discriminative object parts using deep CNN. The fundamental idea is to cluster the last CNN feature maps into groups. The object patches and object parts are obtained from the activations of these clustered feature maps.~\cite{xiao2015application} needs the model to be trained on the category of interest, while we only require the general trained CNN. Similarly, DTRAM~\citep{li2017dynamic} learns to end the attention process for each image after a fixed number of steps. Several methods are proposed to take advantage of object parts. However, the most popular one is the deformable part model (DPM) \citep{felzenszwalb2008DPM}, which learns the constellation relative to the bounding box with Support Vector Machines (SVM). \cite{simon2015neural} improved upon \cite{simon2014part} and employed DPM to localize the parts using the constellation provided by DPM \citep{felzenszwalb2008DPM}. Navigator-Teacher-Scrutinizer Network (NTSNet)~\citep{yang2018learning} uses informative regions in images without employing any annotations. Another teacher-student network was proposed recently as Trilinear Attention Sampling Network (TASN)~\citep{zheng2019TASN}, which is composed of a trilinear attention module, attention-based sampler, and a feature distiller.

\textbf{No Bounding Boxes Methods}. Contrary to utilizing the bounding box annotations, current fine-grain visual categorization state-of-the-art methods avoid incorporating the bounding boxes during testing and training phases altogether.~\cite{zhang2014part} and \cite{lin2015bilinear} used a two-stage network for object and object-part detection and classification employing R-CNN and BilinearCNN, respectively.  Part Stacked CNN~\citep{huang2016part} adopts the same strategy as \cite{lin2015bilinear} and \cite{zhang2014part} of a two-stage system; however, the difference lies in stacking of the object-parts at the end for classification.~\cite{fu2017look} proposed multiple-scale RACNN to acquire distinguishing attention and region feature representations. Moreover, HIHCA~\citep{cai2017higher} incorporated higher-order hierarchical convolutional activations via a kernel scheme.
  
\textbf{Distance metric learning Methods}.  An alternative approach to part-based algorithms is distance learning algorithms which aim to cluster the data points/objects into the same category while moving different types away from each other.~\cite{bromley1994signature} trained Siamese networks using deep metrics for signature verification and, in this context,  set the trend in this direction. Recently, \cite{qian2015fine} employs a multi-stage framework that accepts pre-computed feature maps and learning the distance metric for classification. The pre-computed features can be extracted from DeCAF~\citep{donahue2014decaf}, as these features are discriminative and can be used in many tasks for classification.~\cite{dubey2018pairwise} employs pairwise confusion (PC) via traditional classifiers.

\textbf{Feature Representations Based Methods}. These methods utilize the features from CNN methods to capture the global information. Many works including~\cite{branson2014bird,krause2015fine,xiao2015application}, and \cite{zhang2014part} utilized the feature representations of a CNN and employed in many tasks such as object detection~\citep{girshick2014rich}, Understanding~\citep{zeiler2014visualizing} and recognition~\citep{sharif2014cnn}. CNN captures the global information directly as opposed to the traditional descriptors that capture local information and require manual engineering to encode global representation. Destruction and Construction Learning (DCL)~\citep{chen2019DCL} takes advantage of a standard classification network and emphasizes discriminative local details. The model then reconstructs the semantic correlation among local regions. \cite{zeiler2014visualizing} illustrated the reconstruction of the original image from the activations of the fifth max-pooling layer. Max-pooling ensures invariance to small-scale translation and rotation; however, global spatial information might achieve robustness to larger-scale deformations.~\cite{gong2014multi} combined the features from fully connected layers using VLAD pooling to capture global information. Similarly, \cite{cimpoi2015deep} pooled the features from convolutional layers instead of fully connected layers for text recognition based on the idea that the convolutional layers are transferable and are not domain-specific. Following the footsteps of \cite{cimpoi2015deep}, and \cite{gong2014multi}, PDFR by~\cite{zhang2016picking} encoded the CNN filters responses employing a picking strategy via the combination of Fisher Vectors. However, considering feature encoding as an isolated element is not an optimum choice for convolutional neural networks.

\begin{table}[t]
\caption{Details of six fine-grained visual categorization datasets to evaluate the proposed method.}

\centering
\begin{tabular}{l|rrrr}
\hline
\rowcolor{Gray}&&\multicolumn{3}{c}{No. of Images} \\
\cline{3-5}
\rowcolor{Gray}Dataset   & \multicolumn{1}{c}{Classes} &  \multicolumn{1}{c}{Train}      & \multicolumn{1}{c}{Val}  & \multicolumn{1}{c}{Test}     \\ \hline \hline
NABirds~\citep{van2015NaBirdDataset}   & 555         & 23,929     & \multicolumn{1}{c}{-}         & 24,633   \\
\rowcolor{Gray!50}Dogs~\citep{khosla2011DogDataset}      & 120         & 12,000     & \multicolumn{1}{c}{-}         & 8,580    \\
CUB~\citep{WahCUBDataset}       & 200         & 5,994      & \multicolumn{1}{c}{-}         & 5,794    \\
\rowcolor{Gray!50}Aircraft~\citep{maji2013aircraftdataset}  & 100         & 3,334      & 3,333                         & 3,333    \\
Cars~\citep{krause2013CarDataset}      & 196         & 8,144      & \multicolumn{1}{c}{-}         & 8,041    \\
\rowcolor{Gray!50}Flowers~\citep{nilsback2008FlowerDataset}   & 102         & 2,040      & \multicolumn{1}{c}{-}         & 6,149    \\
\hline \hline
\end{tabular}
\label{tab:FGVC_datasets}
\end{table}

\textbf{Feature Integration Algorithms}. Recently,  feature integration methods take the features from different layers of the same CNN model and combine them. This technique is becoming popular and is adopted by several approaches.  The intuition behind feature integration is to take advantage of global semantic information captured by fully connected layers and instance level information preserved by convolutional layers \citep{babenko2015aggregating}. \cite{long2015fully} merged the features from intermediate and high-level convolutional activations in their convolutional network to exploit both low-level details and high-level semantics for image segmentation. Similarly, for localization and segmentation, \cite{hariharan2015hypercolumns} concatenated the feature maps of convolutional layers at a pixel as a vector to form a descriptor. Likewise, for edge detection, \cite{xie2015holistically} added several feature maps from the lower convolutional layers to guide CNN and predict edges at different scales.

\section{Traditional Networks}
\label{sec:cnn}
To make the paper self-inclusive, we briefly provide the basic building blocks of the modern state-of-the-art traditional CNN architectures. These architectures can be broadly categorized into plain, residual, densely connected, inception, and split-attention networks. We review the most prominent and pioneering traditional networks which fall in each mentioned category and then adapt these models for the fine-grained classification task.  The five architectures we investigate are VGG~\citep{simonyan2014very}, ResNet~\citep{he2016ResNet}, DenseNet~\citep{huang2017denseNet}, Inception~\citep{szegedy2016rethinking}, and ResNest~\citep{zhang2020resnest}. 

\begin{figure}[t]
\centering
\begin{tabular}{c}
\includegraphics[width=0.5\columnwidth,valign=t]{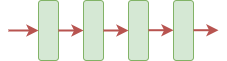}
\end{tabular}
\caption{Basic building block of the VGG~\citep{simonyan2014very}, where no skip connections are used.}
\label{fig:VGG}
\end{figure}

\subsection{Plain Network} Pioneering CNN architectures such as VGG~\citep{simonyan2014very} and AlexNet follow a single path \ie without any skip connections.  The success of AlexNet \citep{krizhevsky2012Alexnet}, inspired VGG. These networks rely on the smaller convolutional filters because a sequence of smaller convolutional filters achieves the same performance compared to a single larger convolutional filter. For example, when four convolutional layers of 3$\times$3 are stacked together, it has the same receptive field as two 5$\times$5 convolutional layers in sequence. Although, the large receptive field has fewer parameters than the smaller ones. The basic building block of VGG~\citep{simonyan2014very} architecture is shown in Figure~\ref{fig:VGG}.

VGG~\citep{simonyan2014very} has many variants; we use the 19-layer convolutional network, which has shown promising results on ImageNet. As mentioned earlier, the block structure of VGG is planar (without any skip connection), and the number of feature channels is increased from 64 to 512.

\subsection{Residual Network}
\begin{figure}[t]
\centering
\begin{tabular}{c}
\includegraphics[width=0.5\columnwidth,valign=t]{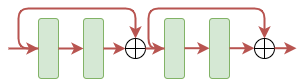}
\end{tabular}
\caption{ResNet~\citep{he2016ResNet}  utilize skip connections inside each module.}
\label{fig:ResNet}
\end{figure}

To solve the vanishing gradients problem, the residual network employed elements of the network with skip connections known as identity shortcuts, as shown in Figure~\ref{fig:ResNet}. The pioneering research in this direction is ResNet~\citep{he2016ResNet}. 

The identity shortcuts help to propagate the gradient signal back without being diminished. The identity shortcuts theoretically \enquote{skip} over all layers and reach the network's initial layers, learning the task at hand. Because of the summation of features at the end of each module, ResNet~\citep{he2016ResNet} learns only an offset, and therefore, it does not require the learning of the full features. The identity shortcuts allow for successful and robust training of much deeper architectures than previously possible.  We compare ResNet50 \& ResNet152 variants with fine-grained classifiers due to successful classification results.

\subsection{Dense Network}

\begin{figure}[t]
\centering
\begin{tabular}{c}
\includegraphics[width=0.5\columnwidth,valign=t]{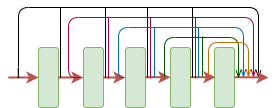}
\end{tabular}
\caption{Basic block of the DenseNet~\citep{huang2017denseNet}, each layer get connection from previous layers of the block.}
\label{fig:Dense}
\end{figure}

Building upon the success of ResNet~\citep{he2016ResNet}, DenseNet~\citep{huang2017denseNet} concatenates each convolutional layer in the modules,  replacing the expensive element-wise addition and retaining the current features and from the previous layers through skipped connections.  Furthermore, there is always a path for information from the last layer backward to deal with the gradient diminishing problem. Moreover, to improve computational efficiency, DenseNet~\citep{huang2017denseNet} utilizes 1$\times$1 convolutional layers to reduce the number of input feature maps before each 3$\times$3 convolutional layer.  Transition layers are applied to compress the number of channels that result from the concatenation operations.  The building block of DenseNet~\citep{huang2017denseNet} is shown in Figure~\ref{fig:Dense}. 

The performance of DenseNet on ILSVRC is comparable with ResNet. However, it has significantly fewer parameters, thus requiring fewer computations, \eg DenseNet with 201 convolutional layers with 20 million parameters produces comparable validation error as a ResNet with 101 convolutional layers having 40 million parameters.  Therefore,  we consider DenseNet, a suitable candidate for fine-grained classification.

\subsection{Inception Network}
\begin{figure}[t]
\centering
\begin{tabular}{c}
\includegraphics[width=0.3\columnwidth,valign=t]{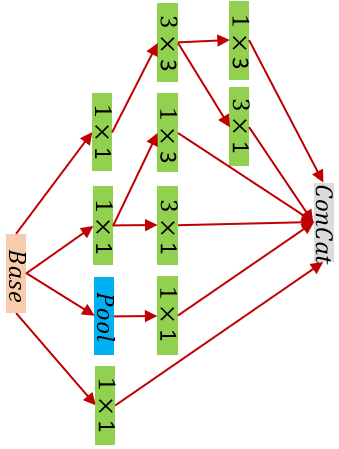}
\end{tabular}
\caption{Basic building block of the Inception-v3~\citep{szegedy2016rethinking}, where many paths are used for feature extraction and then concatenated.}
\label{fig:inception}
\end{figure}

Here, we present Inception-v3~\citep{szegedy2016rethinking}, which utilize label smoothing as a regularization with 7$\times$7 convolutions factorization. Similarly, to propagate label information in the deepest parts of the network, Inception-v3~\citep{szegedy2016rethinking} employ an auxiliary classifier along with batch normalization help for sidehead layers. Figure~\ref{fig:inception} shows the proposed block in the Inception-v3~\citep{szegedy2016rethinking} architecture is used on 8$\times$8 grids of the coarsest level to promote high-dimensional representations.

\subsection{Split-Attention Network}
\begin{figure}[t]
\centering
\begin{tabular}{c}
\includegraphics[width=0.6\columnwidth,valign=t]{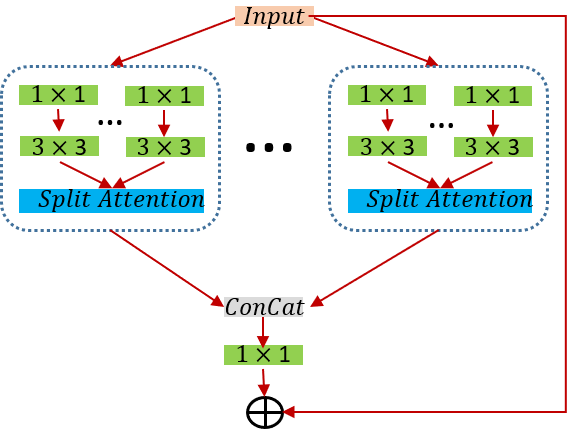}
\end{tabular}
\caption{Basic building block of the ResNest~\citep{zhang2020resnest}, where different paths are used for feature extraction and then concatenated.}
\label{fig:resnest}
\end{figure}

Lastly, we present the split-attention network in Figure~\ref{fig:resnest}, which employs attention and residual block, called ResNest~\citep{zhang2020resnest}, an extension of the resnet. The cardinal group representations are then concatenated along the channel dimension. The final output of other split-attention blocks is produced using a shortcut connection similar to standard residual blocks considering the input and output feature-map have the same shape. Moreover, to align the outputs of blocks having a stride, an appropriate transformation is implemented to the shortcut connection, \eg transformation can be convolution, strided-convolution, or convolution with pooling.

\section{Experiments}
\label{sec:experiments}
\subsection{Experimental Settings}
Stochastic Gradient Descent (SGD)~\cite{bottou2010SGD} optimizer and a decay rate of 1e$^{-4}$ are used. We choose the batch size to be 32,  an initial learning rate of 0.01 for 200 epochs, where the learning rate is decreased linearly by 0.1 after every 30 epochs for all datasets. The networks are finetuned from ImageNet~\citep{deng2009imagenet} training weights.  According to each dataset, the last fully-connected layer is also re-mapped from 1k to the number of classes.

\subsection{Datasets}
This section provides the details of the six most prominent fine-grain datasets used for evaluation and comparison against the current state-of-the-art algorithms.

\begin{itemize}
 \item \textbf{Birds:} The birds' datasets that we compare on are Caltech-UCSD Birds-200-2011, abbreviated as CUB~\citep{WahCUBDataset} is composed of 11,788 photographs of 200 categories which further divided into 5,994 training and 5,794 testing images. The second dataset for birds' fine-grained classification is North American Birds, generally known as NABirds~\citep{van2015NaBirdDataset}, which is the largest in this comparison.  NABirds~\citep{van2015NaBirdDataset} has 555 species found in North America with 48562 photographs.

\item \textbf{Dogs:}  The Stanford Dogs~\citep{khosla2011DogDataset} is a subset of ImageNet~\citep{deng2009imagenet} gathered for the task of fine-grained categorization. The dataset composed of 12k training and 8,580 testing images. 

\item \textbf{Cars:} The cars dataset~\citep{krause2013CarDataset} has 196 classes with different make, model, and year. It has a total number of 16185 car photographs where the split is  8,144 training images and 8,041 testing images \ie roughly 50\% for both. 

\item \textbf{Aeroplanes:} A total of 10,200 images with 102 variants having 100 images for each are present in the fine-grained visual classification of Aircraft \ie FGVC-aircraft dataset~\citep{maji2013aircraftdataset}.  Airplanes are an alternative to objects considered for fine-grained categorization, such as birds and pets.

\item \textbf{Flowers:}  The number of classes in the flower~\citep{nilsback2008FlowerDataset} dataset is 102. The training images are 2,040, while the testing images are 6,149. Furthermore, there are significant variations within categories while having similarities with other categories.
    
\end{itemize}

Table \ref{tab:FGVC_datasets} summarizes the number of classes and the number of images, including the data split for training, testing, and validation (if any) for the fine-grain visualization datasets. 

\section{Evaluations}
\label{sec:evaluations}

\begin{table}[t]
\caption{Comparison of the state-of-the-art fine grain classification on CUB~\citep{WahCUBDataset} dataset}
\centering
\begin{tabular}{cl|c}
\hline
\rowcolor{Gray} CNN&Methods                                & Acc. \\\hline\hline
\multirow{16}{*}{\rotatebox{90}{Fine-Grained}}&  MGCNN~\citep{wang2015multiple}          & 81.7\%\\ 
&\cellcolor[HTML]{EFEFEF}STCNN~\citep{jaderberg2015spatial}      &\cellcolor[HTML]{EFEFEF}84.1\%\\
&FCAN~\citep{liu2016fully}               & 84.3\%\\
& \cellcolor[HTML]{EFEFEF}PDFR~\citep{zhang2016picking}           &\cellcolor[HTML]{EFEFEF}84.5\%\\
&RACNN~\citep{fu2017look}                & 85.3\%\\
& \cellcolor[HTML]{EFEFEF}HIHCA~\citep{cai2017higher}             &\cellcolor[HTML]{EFEFEF}85.3\%\\
&BoostCNN~\citep{moghimi2016boosted}     & 85.6\%\\
&\cellcolor[HTML]{EFEFEF}DTRAM~\citep{li2017dynamic}             &\cellcolor[HTML]{EFEFEF}86.0\%\\
&BilinearCNN~\citep{lin2015bilinear}&84.1\%\\
&\cellcolor[HTML]{EFEFEF}PC-BilinearCNN~\citep{dubey2018pairwise}& \cellcolor[HTML]{EFEFEF}85.6\%\\
&PC-DenseCNN~\citep{dubey2018pairwise}&86.7\%\\
&\cellcolor[HTML]{EFEFEF}\citep{cui2017kernel}         & \cellcolor[HTML]{EFEFEF}86.2\%\\
&MACNN~\citep{zheng2017learning}         &86.5\%\\
&\cellcolor[HTML]{EFEFEF}NTSNet~\citep{yang2018learning}         &\cellcolor[HTML]{EFEFEF}87.5\%\\
&DCL-VGG16~\citep{chen2019DCL}           &86.9\%\\
&\cellcolor[HTML]{EFEFEF}DCL~ResNet50\citep{chen2019DCL}         &\cellcolor[HTML]{EFEFEF}\textbf{87.8}\%\\
&TASN~\citep{zheng2019TASN}              &\textbf{87.9}\%\\ \hline \hline
\multirow{9}{*}{\rotatebox{90}{Traditional}}&\cellcolor[HTML]{EFEFEF}VGG19~\citep{simonyan2014very}          &\cellcolor[HTML]{EFEFEF}77.8\%\\
&ResNet50~\citep{he2016ResNet}           &84.7\% \\ 
&\cellcolor[HTML]{EFEFEF}ResNet152~\citep{he2016ResNet}          &\cellcolor[HTML]{EFEFEF}85.0\% \\ 
&Inception-v3~\citep{szegedy2016rethinking} &76.2\% \\
&\cellcolor[HTML]{EFEFEF}NasNet~\citep{zoph2018learning}         &\cellcolor[HTML]{EFEFEF}83.0\% \\ 
&ResNest50~\citep{zhang2020resnest}&82.3\%\\
&\cellcolor[HTML]{EFEFEF}EfficientNet-B0~\citep{tan2019efficientnet}&\cellcolor[HTML]{EFEFEF}78.0\%\\
&EfficientNet-B4~\citep{tan2019efficientnet} &84.7\%\\
&\cellcolor[HTML]{EFEFEF}EfficientNet-B7~\citep{tan2019efficientnet} &\cellcolor[HTML]{EFEFEF}85.6\%\\
&DenseNet161~\citep{huang2017denseNet}   &\textbf{87.7}\% \\ \hline \hline
\end{tabular}
\label{tab:CUB_dataset}
\end{table}

\subsection{Performance on CUB Dataset}
We present the comparisons on the CUB dataset~\citep{WahCUBDataset} in Table~\ref{tab:CUB_dataset}. The best performer on this dataset is DenseNet, which is unsurprising because the model concatenates the feature maps from preceding layers to preserve details. The worst performing among the traditional classifiers is NasNet~\citep{zoph2018learning}, maybe due to its design, which is more inclined towards a specific dataset (\ie ImageNet~\citep{deng2009imagenet}). The ResNet models perform relatively better than NasNet, which shows that networks with shortcut connections surpass in performance than those with multi-scale representations for fine-grained classification. DenseNet offers higher accuracy than ResNet because the former does not fuse the feature and carry the details forward, unlike the latter, where the features are combined in each block. 

\begin{table}[t]
\caption{Experimental results on FGVC Aircraft~\citep{maji2013aircraftdataset} and Cars~\citep{krause2013CarDataset}.}
\centering
\begin{tabular}{cl|cc}
\hline
\rowcolor{Gray}               &             &\multicolumn{2}{c}{Datasets}      \\\cline{3-4}
\rowcolor{Gray} CNN&\multicolumn{1}{c|}{Methods} & Aircraft  & Cars \\  \hline \hline 
\multirow{13}{*}{\rotatebox{90}{Fine-Grained}}&FVCNN~\citep{gosselin2014revisiting}     &81.5\%    &-\\
&\cellcolor[HTML]{EFEFEF}FCAN~\citep{liu2016fully}                & \cellcolor[HTML]{EFEFEF}-         &\cellcolor[HTML]{EFEFEF}89.1\%\\
&BilinearCNN~\citep{lin2015bilinear}      &84.1\%    &91.3\%\\
&\cellcolor[HTML]{EFEFEF}RACNN~\citep{fu2017look}                 &\cellcolor[HTML]{EFEFEF}88.2\%    &\cellcolor[HTML]{EFEFEF}92.5\%\\
&HIHCA~\citep{cai2017higher}              &88.3\%    &91.7\%\\
&\cellcolor[HTML]{EFEFEF}BoostCNN~\citep{moghimi2016boosted}      &\cellcolor[HTML]{EFEFEF}88.5\%    &\cellcolor[HTML]{EFEFEF}92.1\%\\
&\cite{cui2017kernel}          &88.5\%    &92.4\%\\
&\cellcolor[HTML]{EFEFEF}PC-BilinearCNN~\citep{dubey2018pairwise}    &\cellcolor[HTML]{EFEFEF}85.8\%    &\cellcolor[HTML]{EFEFEF}92.5\%\\
&PC-ResCNN~\citep{dubey2018pairwise}	&83.4\%    &93.4\%\\
&\cellcolor[HTML]{EFEFEF}PC-DenseCNN~\citep{dubey2018pairwise} &\cellcolor[HTML]{EFEFEF}89.2\%    &\cellcolor[HTML]{EFEFEF}92.7\%\\
&MACNN~\citep{zheng2017learning}          &89.9\%    &92.8\%\\
&\cellcolor[HTML]{EFEFEF}DTRAM~\citep{li2017dynamic}              &\cellcolor[HTML]{EFEFEF}-         &\cellcolor[HTML]{EFEFEF}93.1\%\\
&TASN~\citep{zheng2019TASN}&-          &93.8\%\\
&\cellcolor[HTML]{EFEFEF}NTSNet~\citep{yang2018learning}          &\cellcolor[HTML]{EFEFEF}91.4\%    &\cellcolor[HTML]{EFEFEF}93.9\%\\\hline \hline
\multirow{9}{*}{\rotatebox{90}{Traditional}}&VGG19~\citep{simonyan2014very}          & \cellcolor[HTML]{EFEFEF}85.7\%  & \cellcolor[HTML]{EFEFEF}80.5\%\\
&\cellcolor[HTML]{EFEFEF}ResNet50~\citep{he2016ResNet}            &\cellcolor[HTML]{EFEFEF}91.4\%    &\cellcolor[HTML]{EFEFEF}91.7\%\\ 
&ResNet152~\citep{he2016ResNet}           &90.7\%    &93.2\%\\ 
&\cellcolor[HTML]{EFEFEF}NasNet~\citep{zoph2018learning}          &\cellcolor[HTML]{EFEFEF}88.5\%    &\cellcolor[HTML]{EFEFEF}-\\
&Inception-v3~\citep{szegedy2016rethinking} & 85.4\%  & 85.8\%   \\
&\cellcolor[HTML]{EFEFEF}ResNest50~\citep{zhang2020resnest}&\cellcolor[HTML]{EFEFEF}89.9\%&\cellcolor[HTML]{EFEFEF}89.6\%\\
&EfficientNet-B0~\citep{tan2019efficientnet}&80.9\% &82.8\% \\
&\cellcolor[HTML]{EFEFEF}EfficientNet-B4~\citep{tan2019efficientnet} &\cellcolor[HTML]{EFEFEF}86.8\% &\cellcolor[HTML]{EFEFEF}86.9\% \\
&EfficientNet-B7~\citep{tan2019efficientnet} &92.0\% &90.2\% \\
&\cellcolor[HTML]{EFEFEF}DenseNet161~\citep{huang2017denseNet}    &\cellcolor[HTML]{EFEFEF}\textbf{92.9}\%    &\cellcolor[HTML]{EFEFEF}\textbf{94.5}\%\\ \hline \hline
\end{tabular}
\label{tab:Aircraft_cars_comparisons}
\end{table}

The fine-grained classification literature considers CUB-200-2011~\citep{WahCUBDataset} as a standard benchmark for evaluation; therefore, image-level labels, bounding boxes, and different types of annotations are employed to extract the best results on this dataset. Similarly, multi-branch networks focusing on various parts of images and multiple objective functions are combined for optimization. On the contrary, the traditional classifiers~\citep{huang2017denseNet,he2016ResNet} use a single loss without any extra information or any other annotations. The best-performing fine-grained classifiers for CUB~\citep{WahCUBDataset} are DCL~ResNet50~\citep{chen2019DCL}, TASN~\citep{zheng2019TASN}, and NTSNet~\citep{yang2018learning} where merely 0.1\% and 0.2\% gain is recorded over DenseNet~\citep{huang2017denseNet} for~\citep{chen2019DCL} and~\citep{zheng2019TASN}, respectively. Furthermore, NTSNet~\citep{yang2018learning} lags by a margin of 0.2\%. The improvement over DenseNet is insignificant, keeping in mind the different computationally expensive tactics employed to learn the distinguishable features by fine-grained classifiers.

\begin{table*}[t]
\caption{Comparison of the state-of-the-art fine grain classification on Dogs~\citep{khosla2011DogDataset}, Flowers~\citep{nilsback2008FlowerDataset} and NABirds~\citep{van2015NaBirdDataset} dataset.}
\centering
\begin{tabular}{cl|ccc}
\hline
 \rowcolor{Gray}             &           &\multicolumn{3}{c}{Datasets}      \\\cline{3-5}
\rowcolor{Gray}CNN&\multicolumn{1}{c|}{Methods} &Dogs   &Flowers   &NABirds\\ \hline \hline
\multirow{10}{*}{\rotatebox{90}{Fine-Grained}} &\cite{zhang2016weakly}          &80.4\% & -       & -\\
&\cellcolor[HTML]{EFEFEF}\cite{krause2016unreasonable}  &\cellcolor[HTML]{EFEFEF}80.6\% & \cellcolor[HTML]{EFEFEF}-       & \cellcolor[HTML]{EFEFEF}-\\ 
&Det.+Seg.~\citep{angelova2013efficient}      & -     & 80.7\%  & -\\
&\cellcolor[HTML]{EFEFEF}Overfeat~\citep{sharif2014cnn}               &\cellcolor[HTML]{EFEFEF} -     & \cellcolor[HTML]{EFEFEF}86.8\%  &\cellcolor[HTML]{EFEFEF}-\\
&\cite{branson2014bird}        & -     & -       & 35.7\% \\ 
&\cellcolor[HTML]{EFEFEF}\cite{van2015NaBirdDataset}       &\cellcolor[HTML]{EFEFEF}-     &\cellcolor[HTML]{EFEFEF}-       & \cellcolor[HTML]{EFEFEF}75.0\% \\ 
&BilinearCNN~\citep{lin2015bilinear}          &82.1\% & 92.5\%  & 80.9\% \\
&\cellcolor[HTML]{EFEFEF}PC-ResCNN~\citep{dubey2018pairwise}        &\cellcolor[HTML]{EFEFEF}73.4\% & \cellcolor[HTML]{EFEFEF}93.5\%  &\cellcolor[HTML]{EFEFEF}68.2\%\\
&PC-BilinearCNN~\citep{dubey2018pairwise}     &83.0\% & 93.7\%  & 82.0\%\\
&\cellcolor[HTML]{EFEFEF}PC-DenseCNN~\citep{dubey2018pairwise}     &\cellcolor[HTML]{EFEFEF}83.8\% & \cellcolor[HTML]{EFEFEF}91.4\%  & \cellcolor[HTML]{EFEFEF}82.8\%\\ \hline\hline
\multirow{8}{*}{\rotatebox{90}{Traditional}}&VGG19~\citep{simonyan2014very}               & 76.7\%     & 88.73\% & 74.9\%\\
&\cellcolor[HTML]{EFEFEF}ResNet50~\citep{he2016ResNet}                &\cellcolor[HTML]{EFEFEF}83.4\% & \cellcolor[HTML]{EFEFEF}97.2\%  & \cellcolor[HTML]{EFEFEF}79.6\%\\
&ResNet152~\citep{he2016ResNet}               &85.2\% & 97.5\%  & 84.0\%\\
&\cellcolor[HTML]{EFEFEF} Inception-v3~\citep{szegedy2016rethinking} & \cellcolor[HTML]{EFEFEF}85.8\% &\cellcolor[HTML]{EFEFEF}93.3\% & \cellcolor[HTML]{EFEFEF}68.4\% \\
&ResNest50~\citep{zhang2020resnest}&87.7\%&94.7\%&80.4\%\\
&\cellcolor[HTML]{EFEFEF}EfficientNet-B0~\citep{tan2019efficientnet}&\cellcolor[HTML]{EFEFEF}84.9\% &\cellcolor[HTML]{EFEFEF}91.4\% & \cellcolor[HTML]{EFEFEF}63.7\%\\
&EfficientNet-B4~\citep{tan2019efficientnet} &92.4\% &92.8\% &77.0\%\\
&\cellcolor[HTML]{EFEFEF}EfficientNet-B7~\citep{tan2019efficientnet} &\cellcolor[HTML]{EFEFEF}\textbf{93.6}\% &\cellcolor[HTML]{EFEFEF}96.2\% &\cellcolor[HTML]{EFEFEF}- \\
&DenseNet161~\citep{huang2017denseNet}        &85.2\% &\textbf{98.1\%} &\textbf{86.3}\%\\ \hline \hline
\end{tabular}
\label{tab:three_dataset}
\end{table*}

\subsection{Quantitative analysis on Aircraft and Cars}
 In Table~\ref{tab:Aircraft_cars_comparisons}, the performances of fine-grained classifiers are shown on Cars and Aircraft datasets. Here, we also observe that the performance of the traditional classifiers is better than the fine-grained classifiers. DenseNet161 has an improvement of about 1.5\% and 3\% on Aircraft~\citep{maji2013aircraftdataset} compared to best-performing NTSNet~\citep{yang2018learning} and MACNN~\citep{zheng2017learning}, respectively. Similarly, an improvement of 0.6\% and 1.4\% is recorded against NTSNet~\citep{yang2018learning} and DTRAM~\citep{li2017dynamic} on the Cars dataset, respectively. The fine-grain classifiers such as~\cite{yang2018learning,li2017dynamic} and \cite{zheng2019TASN} fail to achieve the same accuracy as the traditional classifiers, although the former employ more image-specific information for learning.

\subsection{Comparison on Stanford Dogs} 
The Stanford dogs~\citep{khosla2011DogDataset} is another challenging dataset where the performance is compared in Table~\ref{tab:three_dataset}. Here, we utilize ResNet and DenseNet from the traditional ones. The performance of ResNet composed of 152 layers is similar to DenseNet with 161 layers; both achieved 85.2\% accuracy, which is 1.4\% higher than PC-DenseCNN~\citep{dubey2018pairwise}, the best performing method in fine-grained classifiers. This experiment suggests that incorporating traditional classifiers in the fine-grained ones requires more insight than just utilizing them in the framework. It is also worth mentioning that some of the fine-grained classifiers employ a large amount of data from other sources in addition to the Stanford dogs' training data.

\subsection{Results of Flower dataset}
The accuracy of DenseNet on the Flower dataset~\citep{nilsback2008FlowerDataset} is 98.1\% which is  around 5.5\% higher as compared to the second-best performing state-of-the-art method (PC-ResCNN~\citep{dubey2018pairwise}) in Table~~\ref{tab:three_dataset}. Similarly, the other traditional classifiers also outperform the fine-grained ones by a significant margin. It should also be noted that the performance on this dataset is approaching saturation.

\subsection{Performance on NABirds}
Relatively fewer methods have reported their results on this dataset. However, for the sake of completeness, we provide comparisons on the NABirds~\citep{van2015NaBirdDataset} dataset. Again the leading performance on NABirds is achieved by DenseNet161, followed by ResNet152.  The third-best performer is a fine-grain classifier \ie PC-DenseCNN~\citep{dubey2018pairwise}, which internally employes DenseNet161 lags behind by 3.5\%. This shows the superior performance of the traditional CNN classifiers against state-of-the-art fine-grained CNN classifiers.

\subsection{Ablation studies}

\noindent
\textbf{Fine-tune vs. Scratch}: Here, we present two strategies for training traditional CNN classification networks \ie fine-tuning the weights via ImageNet~\citep{deng2009imagenet} and training from scratch (randomly initializing the weights) for the Car dataset. The accuracy presented for each is given in Table~\ref{tab:trainingScratch}. The ResNet50 achieves higher accuracy when fine-tuned as compared to the randomly initialized version. Similarly, ResNet152 performed better for the fine-tuned network; however, it fails when trained from scratch. The reason may be due to a large number of parameters and smaller training data.

\begin{table}[t]
\caption{Differences strategies for initialing the network weights \ie fine-tuning from ImageNet and random initialization (scratch) for
 Cars~\citep{krause2013CarDataset} dataset.}
\centering
\begin{tabular}{l|cc}
\hline
\multicolumn{1}{c|}{Initial} &\multicolumn{2}{c}{Methods}      \\\cline{2-3}
\multicolumn{1}{c|}{Weights} &ResNet50 & ResNet152\\  \hline \hline 
Scratch   &83.4\%    &36.9\%\\ 
Fine-tune &91.7\%    &93.2\%\\ 
\hline \hline
\end{tabular}
\label{tab:trainingScratch}
\end{table}

\vspace{1mm}
\noindent
\textbf{Backbones Improvement Over Standalone Classifiers}:
Some fine-grain state-of-the-art methods use ResNet50 as the backbone and achieve an accuracy higher than the standalone ResNet50. To be precise, Table~\ref{tab:backbones} shows the backbones used by state-of-the-art methods in their algorithm.  One can observe that many algorithms employ the same backbones more than once, increasing the overhead and doubling or tripling the number of parameters. Besides utilizing traditional classifiers as backbones, state-of-the-art fine-grain methods rely on specialized techniques to extract fine-grain features; hence, adding more parameters and computation. Therefore, the improvement achieved by the state-of-the-art fine-grain methods comes at the cost of extra considerations and the number of parameters, while the traditional classifier like DenseNet doesn't require such tricks to achieve the same accuracy.

\begin{table}[t]
\caption{The comparison of backbone and number of parameters in fine-grain methods regarding classification accuracy on the CUB dataset. The input to all methods is 448 $\times$ 448.}
\centering
\begin{tabular}{l|ccc}
\hline
\rowcolor{Gray} Methods                                &Backbone              &Parameters &Accuracy\\ \hline \hline
MGCNN~\citep{wang2015multiple}                         &3$\times$VGG16        &429M   &81.7\\
\rowcolor{Gray!50}STCNN~\citep{jaderberg2015spatial}   &3$\times$Inception-v2 &71.5M  &84.1\\
RA-CNN~\citep{cai2017higher}                           &3$\times$VGG19        &429M   &85.3\\
\rowcolor{Gray!50}MACNN~\citep{zheng2017learning}      &3$\times$VGG19        &144M   &85.4\\
TASN~\citep{zheng2019TASN}                             &1$\times$VGG19        &140M   &87.1\\
\rowcolor{Gray!50}MAMC~\citep{sun2018multi}            &1$\times$Resnet50     &434M   &86.5\\
NTSNet~\citep{yang2018learning}                        &3$\times$Resnet50     &25.5M  &87.3\\
\rowcolor{Gray!50}TASN~\citep{zheng2019TASN}           &1$\times$Resnet50     &35.2M  &87.9\\ \hline
DenseNet~\citep{huang2017denseNet}                     &1$\times$DenseNet161  &28.7M  &87.7\\ \hline \hline
\end{tabular}
\label{tab:backbones}
\end{table}

\vspace{1mm}
\noindent
\textbf{Parameters, FLOPs, and Performance}: We provide comparisons in terms of the number of parameters, FLOPs, and performance on the ImageNet for the traditional classifiers that have been employed in our experiments in Table~\ref{tab:pp}. The ResNet50~\citep{he2016ResNet} approximately has the same number of parameters as DenseNet161~\citep{huang2017denseNet}, but the performance of DenseNet161~\citep{huang2017denseNet} is much higher than ResNet50~\citep{he2016ResNet}. It should also be noted that DenseNet169 and DenseNet201 have fewer parameters but higher performance on imageNet; hence, we argue that backbones in the fine-grain methods should be updated to appropriate ones as suggested by our experimental analysis.

\begin{table}[t]
\caption{Traditional classifiers comparison on ImageNet~\citep{deng2009imagenet} in terms of number of parameters, FLOPS and accuracy.}
\centering
\begin{tabular}{l|llcc}
\hline
\rowcolor{Gray}              & No. of    &           & \multicolumn{2}{c}{Accuracy}   \\\cline{4-5}
\rowcolor{Gray}Methods       &Parameters &FLOPS      &Top-1      &Top-5 \\ \hline \hline
ResNet18      &11.69M     &1819.06M   & 69.76\%   &89.08\% \\
\rowcolor{Gray!50}ResNet34	  &21.97M     &3671.26M   &73.3\%	  &91.42\% \\
ResNet50	  &25.60M     &4111.51M   &76.15\%	  &92.87\%\\
\rowcolor{Gray!50}ResNet101	  &44.60M     &7833.97M   &77.37\%	  &93.56\%\\
ResNet152	  &60.20M     &11558.83M  &78.31\%	  &94.06\%\\
\rowcolor{Gray!50}Densenet121   &7.98M      &2865.67M   &74.65\%	  &92.17\%\\
Densenet161   &28.68M     &7787.01M   &77.65\%	  &93.80\%\\
\rowcolor{Gray!50}Densenet169   &14.15M     &3398.07M   &76.00\%	  &93.00\%\\
Densenet201   &20.01M     &4340.97M   &77.20\%    &93.57\%\\
\rowcolor{Gray!50}Inception-v3  &23.83M     &5731.28M   &77.45\%	  &93.56\%\\ \hline \hline
\end{tabular}
\label{tab:pp}
\end{table}

\section{Conclusion}
\label{sec:conclusion}
In this paper, we provided comparisons between state-of-the-art traditional CNN classifiers and fine-grained CNN classifiers.  It has been shown that conventional models achieve state-of-the-art performance on fine-grained classification datasets and outperform the fine-grained CNN classifiers. Therefore, it is necessary to update the baselines for comparisons.  It is also important to note that the performance increase is due to the initial weights trained on the ImageNet~\citep{deng2009imagenet} datasets. Furthermore, we have established that the DenseNet161 model achieves new state-of-the-art results for all datasets outperforming the fine-grained classifiers by a significant margin.

\bibliographystyle{spbasic_updated}     
\bibliography{refs}   

\begin{thebibliography}{75}
\providecommand{\natexlab}[1]{#1}
\providecommand{\url}[1]{{#1}}
\providecommand{\urlprefix}{URL }
\expandafter\ifx\csname urlstyle\endcsname\relax
  \providecommand{\doi}[1]{DOI~\discretionary{}{}{}#1}\else
  \providecommand{\doi}{DOI~\discretionary{}{}{}\begingroup
  \urlstyle{rm}\Url}\fi
\providecommand{\eprint}[2][]{\url{#2}}

\bibitem[{Aafaq et~al.(2019)Aafaq, Mian, Liu, Gilani, and
  Shah}]{aafaq2019video}
Aafaq N, Mian A, Liu W, Gilani SZ, Shah M (2019) Video description: A survey of
  methods, datasets, and evaluation metrics. ACM Computing Surveys (CSUR)

\bibitem[{Angelova and Zhu(2013)}]{angelova2013efficient}
Angelova A, Zhu S (2013) Efficient object detection and segmentation for
  fine-grained recognition. In: IEEE/CVF Conference on Computer Vision and
  Pattern Recognition (CVPR)

\bibitem[{Angelova et~al.(2013)Angelova, Zhu, and Lin}]{angelova2013image}
Angelova A, Zhu S, Lin Y (2013) Image segmentation for large-scale subcategory
  flower recognition. In: IEEE Winter Conference on Applications of Computer
  Vision (WACV)

\bibitem[{Babenko and Lempitsky(2015)}]{babenko2015aggregating}
Babenko A, Lempitsky V (2015) Aggregating local deep features for image
  retrieval. In: Proceedings of the IEEE/CVF International Conference on
  Computer Vision (ICCV)

\bibitem[{Berg and Belhumeur(2013)}]{berg2013poof}
Berg T, Belhumeur P (2013) Poof: Part-based one-vs.-one features for
  fine-grained categorization, face verification, and attribute estimation. In:
  IEEE/CVF Conference on Computer Vision and Pattern Recognition (CVPR)

\bibitem[{Bottou(2010)}]{bottou2010SGD}
Bottou L (2010) Large-scale machine learning with stochastic gradient descent.
  In: Proceedings of COMPSTAT, Springer, pp 177--186

\bibitem[{Branson et~al.(2014)Branson, Van~Horn, Belongie, and
  Perona}]{branson2014bird}
Branson S, Van~Horn G, Belongie S, Perona P (2014) Bird species categorization
  using pose normalized deep convolutional nets. arXiv

\bibitem[{Bromley et~al.(1994)Bromley, Guyon, LeCun, S{\"a}ckinger, and
  Shah}]{bromley1994signature}
Bromley J, Guyon I, LeCun Y, S{\"a}ckinger E, Shah R (1994) Signature
  verification using a" siamese" time delay neural network. In: Advances in
  Neural Information Processing Systems (NIPS)

\bibitem[{Cai et~al.(2017)Cai, Zuo, and Zhang}]{cai2017higher}
Cai S, Zuo W, Zhang L (2017) Higher-order integration of hierarchical
  convolutional activations for fine-grained visual categorization. In:
  Proceedings of the IEEE/CVF International Conference on Computer Vision
  (ICCV)

\bibitem[{Chai et~al.(2012)Chai, Rahtu, Lempitsky, Van~Gool, and
  Zisserman}]{chai2012tricos}
Chai Y, Rahtu E, Lempitsky V, Van~Gool L, Zisserman A (2012) Tricos: A
  tri-level class-discriminative co-segmentation method for image
  classification. In: Proceedings of the European Conference on Computer Vision
  (ECCV)

\bibitem[{Chai et~al.(2013)Chai, Lempitsky, and Zisserman}]{chai2013symbiotic}
Chai Y, Lempitsky V, Zisserman A (2013) Symbiotic segmentation and part
  localization for fine-grained categorization. In: Proceedings of the IEEE/CVF
  International Conference on Computer Vision (ICCV)

\bibitem[{Chen et~al.(2009)Chen, Dhingra, Wu, Yang, Sukthankar, and
  Yang}]{chen2009pfid}
Chen M, Dhingra K, Wu W, Yang L, Sukthankar R, Yang J (2009) Pfid: Pittsburgh
  fast-food image dataset. In: IEEE International Conference on Image
  Processing (ICIP)

\bibitem[{Chen et~al.(2019)Chen, Bai, Zhang, and Mei}]{chen2019DCL}
Chen Y, Bai Y, Zhang W, Mei T (2019) Destruction and construction learning for
  fine-grained image recognition. In: IEEE/CVF Conference on Computer Vision
  and Pattern Recognition (CVPR)

\bibitem[{Cimpoi et~al.(2015)Cimpoi, Maji, and Vedaldi}]{cimpoi2015deep}
Cimpoi M, Maji S, Vedaldi A (2015) Deep filter banks for texture recognition
  and segmentation. In: IEEE/CVF Conference on Computer Vision and Pattern
  Recognition (CVPR)

\bibitem[{Cui et~al.(2017)Cui, Zhou, Wang, Liu, Lin, and
  Belongie}]{cui2017kernel}
Cui Y, Zhou F, Wang J, Liu X, Lin Y, Belongie S (2017) Kernel pooling for
  convolutional neural networks. In: IEEE/CVF Conference on Computer Vision and
  Pattern Recognition (CVPR)

\bibitem[{Deng et~al.(2009)Deng, Dong, Socher, Li, Li, and
  Fei-Fei}]{deng2009imagenet}
Deng J, Dong W, Socher R, Li LJ, Li K, Fei-Fei L (2009) Imagenet: A large-scale
  hierarchical image database. In: IEEE/CVF Conference on Computer Vision and
  Pattern Recognition (CVPR)

\bibitem[{Deng et~al.(2013)Deng, Krause, and Fei-Fei}]{deng2013fine}
Deng J, Krause J, Fei-Fei L (2013) Fine-grained crowdsourcing for fine-grained
  recognition. In: IEEE/CVF Conference on Computer Vision and Pattern
  Recognition (CVPR)

\bibitem[{Donahue et~al.(2014)Donahue, Jia, Vinyals, Hoffman, Zhang, Tzeng, and
  Darrell}]{donahue2014decaf}
Donahue J, Jia Y, Vinyals O, Hoffman J, Zhang N, Tzeng E, Darrell T (2014)
  Decaf: A deep convolutional activation feature for generic visual
  recognition. In: International Conference on Machine Learning (ICML)

\bibitem[{Dubey et~al.(2018)Dubey, Gupta, Guo, Raskar, Farrell, and
  Naik}]{dubey2018pairwise}
Dubey A, Gupta O, Guo P, Raskar R, Farrell R, Naik N (2018) Pairwise confusion
  for fine-grained visual classification. In: Proceedings of the European
  Conference on Computer Vision (ECCV)

\bibitem[{Felzenszwalb et~al.(2008)Felzenszwalb, McAllester, and
  Ramanan}]{felzenszwalb2008DPM}
Felzenszwalb P, McAllester D, Ramanan D (2008) A discriminatively trained,
  multiscale, deformable part model. In: IEEE/CVF Conference on Computer Vision
  and Pattern Recognition (CVPR)

\bibitem[{Fu et~al.(2017)Fu, Zheng, and Mei}]{fu2017look}
Fu J, Zheng H, Mei T (2017) Look closer to see better: Recurrent attention
  convolutional neural network for fine-grained image recognition. In: IEEE/CVF
  Conference on Computer Vision and Pattern Recognition (CVPR)

\bibitem[{Gavves et~al.(2013)Gavves, Fernando, Snoek, Smeulders, and
  Tuytelaars}]{gavves2013fine}
Gavves E, Fernando B, Snoek CG, Smeulders AW, Tuytelaars T (2013) Fine-grained
  categorization by alignments. In: Proceedings of the IEEE/CVF International
  Conference on Computer Vision (ICCV)

\bibitem[{Girshick et~al.(2014)Girshick, Donahue, Darrell, and
  Malik}]{girshick2014rich}
Girshick R, Donahue J, Darrell T, Malik J (2014) Rich feature hierarchies for
  accurate object detection and semantic segmentation. In: IEEE/CVF Conference
  on Computer Vision and Pattern Recognition (CVPR)

\bibitem[{Gong et~al.(2014)Gong, Wang, Guo, and Lazebnik}]{gong2014multi}
Gong Y, Wang L, Guo R, Lazebnik S (2014) Multi-scale orderless pooling of deep
  convolutional activation features. In: Proceedings of the European Conference
  on Computer Vision (ECCV)

\bibitem[{Gosselin et~al.(2014)Gosselin, Murray, J{\'e}gou, and
  Perronnin}]{gosselin2014revisiting}
Gosselin PH, Murray N, J{\'e}gou H, Perronnin F (2014) Revisiting the fisher
  vector for fine-grained classification. PRL

\bibitem[{Hariharan et~al.(2015)Hariharan, Arbel{\'a}ez, Girshick, and
  Malik}]{hariharan2015hypercolumns}
Hariharan B, Arbel{\'a}ez P, Girshick R, Malik J (2015) Hypercolumns for object
  segmentation and fine-grained localization. In: IEEE/CVF Conference on
  Computer Vision and Pattern Recognition (CVPR)

\bibitem[{He et~al.(2016)He, Zhang, Ren, and Sun}]{he2016ResNet}
He K, Zhang X, Ren S, Sun J (2016) Deep residual learning for image
  recognition. In: IEEE/CVF Conference on Computer Vision and Pattern
  Recognition (CVPR)

\bibitem[{Hou et~al.(2017)Hou, Feng, and Wang}]{hou2017vegfru}
Hou S, Feng Y, Wang Z (2017) Vegfru: A domain-specific dataset for fine-grained
  visual categorization. In: Proceedings of the IEEE/CVF International
  Conference on Computer Vision (ICCV)

\bibitem[{Huang et~al.(2017)Huang, Liu, Van Der~Maaten, and
  Weinberger}]{huang2017denseNet}
Huang G, Liu Z, Van Der~Maaten L, Weinberger KQ (2017) Densely connected
  convolutional networks. In: IEEE/CVF Conference on Computer Vision and
  Pattern Recognition (CVPR)

\bibitem[{Huang et~al.(2016)Huang, Xu, Tao, and Zhang}]{huang2016part}
Huang S, Xu Z, Tao D, Zhang Y (2016) Part-stacked cnn for fine-grained visual
  categorization. In: IEEE/CVF Conference on Computer Vision and Pattern
  Recognition (CVPR)

\bibitem[{Jaderberg et~al.(2015)Jaderberg, Simonyan, Zisserman
  et~al.}]{jaderberg2015spatial}
Jaderberg M, Simonyan K, Zisserman A, et~al. (2015) Spatial transformer
  networks. In: Advances in Neural Information Processing Systems (NIPS)

\bibitem[{Khosla et~al.(2011)Khosla, Jayadevaprakash, Yao, and
  Li}]{khosla2011DogDataset}
Khosla A, Jayadevaprakash N, Yao B, Li FF (2011) Novel dataset for fgvc:
  Stanford dogs. In: IEEE/CVF Conference on Computer Vision and Pattern
  Recognition (CVPR) Workshop

\bibitem[{Krause et~al.(2013)Krause, Stark, Deng, and
  Fei-Fei}]{krause2013CarDataset}
Krause J, Stark M, Deng J, Fei-Fei L (2013) 3d object representations for
  fine-grained categorization. In: IEEE/CVF Conference on Computer Vision and
  Pattern Recognition (CVPR) Worshops

\bibitem[{Krause et~al.(2015)Krause, Jin, Yang, and Fei-Fei}]{krause2015fine}
Krause J, Jin H, Yang J, Fei-Fei L (2015) Fine-grained recognition without part
  annotations. In: IEEE/CVF Conference on Computer Vision and Pattern
  Recognition (CVPR)

\bibitem[{Krause et~al.(2016)Krause, Sapp, Howard, Zhou, Toshev, Duerig,
  Philbin, and Fei-Fei}]{krause2016unreasonable}
Krause J, Sapp B, Howard A, Zhou H, Toshev A, Duerig T, Philbin J, Fei-Fei L
  (2016) The unreasonable effectiveness of noisy data for fine-grained
  recognition. In: Proceedings of the European Conference on Computer Vision
  (ECCV)

\bibitem[{Krizhevsky et~al.(2012)Krizhevsky, Sutskever, and
  Hinton}]{krizhevsky2012Alexnet}
Krizhevsky A, Sutskever I, Hinton GE (2012) Imagenet classification with deep
  convolutional neural networks. In: Advances in Neural Information Processing
  Systems (NIPS)

\bibitem[{Li et~al.(2017)Li, Yang, Liu, Zhou, Wen, and Xu}]{li2017dynamic}
Li Z, Yang Y, Liu X, Zhou F, Wen S, Xu W (2017) Dynamic computational time for
  visual attention. In: Proceedings of the IEEE/CVF International Conference on
  Computer Vision (ICCV)

\bibitem[{Lin et~al.(2015)Lin, RoyChowdhury, and Maji}]{lin2015bilinear}
Lin TY, RoyChowdhury A, Maji S (2015) Bilinear cnn models for fine-grained
  visual recognition. In: Proceedings of the IEEE/CVF International Conference
  on Computer Vision (ICCV)

\bibitem[{Liu et~al.(2012)Liu, Kanazawa, Jacobs, and Belhumeur}]{liu2012dog}
Liu J, Kanazawa A, Jacobs D, Belhumeur P (2012) Dog breed classification using
  part localization. In: Proceedings of the European Conference on Computer
  Vision (ECCV)

\bibitem[{Liu et~al.(2018)Liu, Qi, Qin, Shi, and Jia}]{liu2018segment}
Liu S, Qi L, Qin H, Shi J, Jia J (2018) Path aggregation network for instance
  segmentation. In: IEEE/CVF Conference on Computer Vision and Pattern
  Recognition (CVPR)

\bibitem[{Liu et~al.(2016)Liu, Xia, Wang, Yang, Zhou, and Lin}]{liu2016fully}
Liu X, Xia T, Wang J, Yang Y, Zhou F, Lin Y (2016) Fully convolutional
  attention networks for fine-grained recognition. arXiv

\bibitem[{Long et~al.(2015)Long, Shelhamer, and Darrell}]{long2015fully}
Long J, Shelhamer E, Darrell T (2015) Fully convolutional networks for semantic
  segmentation. In: IEEE/CVF Conference on Computer Vision and Pattern
  Recognition (CVPR)

\bibitem[{Maji et~al.(2013)Maji, Rahtu, Kannala, Blaschko, and
  Vedaldi}]{maji2013aircraftdataset}
Maji S, Rahtu E, Kannala J, Blaschko M, Vedaldi A (2013) Fine-grained visual
  classification of aircraft. arXiv

\bibitem[{Moghimi et~al.(2016)Moghimi, Belongie, Saberian, Yang, Vasconcelos,
  and Li}]{moghimi2016boosted}
Moghimi M, Belongie SJ, Saberian MJ, Yang J, Vasconcelos N, Li LJ (2016)
  Boosted convolutional neural networks. In: British Machine Vision Conference

\bibitem[{Nilsback and Zisserman(2008)}]{nilsback2008FlowerDataset}
Nilsback ME, Zisserman A (2008) Automated flower classification over a large
  number of classes. In: Ind. Conference on Vision Graphics and Image
  Processing

\bibitem[{Parkhi et~al.(2011)Parkhi, Vedaldi, Jawahar, and
  Zisserman}]{parkhi2011truth}
Parkhi OM, Vedaldi A, Jawahar C, Zisserman A (2011) The truth about cats and
  dogs. In: Proceedings of the IEEE/CVF International Conference on Computer
  Vision (ICCV)

\bibitem[{Parkhi et~al.(2012)Parkhi, Vedaldi, Zisserman, and
  Jawahar}]{parkhi2012cats}
Parkhi OM, Vedaldi A, Zisserman A, Jawahar C (2012) Cats and dogs. In: IEEE/CVF
  Conference on Computer Vision and Pattern Recognition (CVPR)

\bibitem[{Qian et~al.(2015)Qian, Jin, Zhu, and Lin}]{qian2015fine}
Qian Q, Jin R, Zhu S, Lin Y (2015) Fine-grained visual categorization via
  multi-stage metric learning. In: IEEE/CVF Conference on Computer Vision and
  Pattern Recognition (CVPR)

\bibitem[{Sharif~Razavian et~al.(2014)Sharif~Razavian, Azizpour, Sullivan, and
  Carlsson}]{sharif2014cnn}
Sharif~Razavian A, Azizpour H, Sullivan J, Carlsson S (2014) Cnn features
  off-the-shelf: an astounding baseline for recognition. In: IEEE/CVF
  Conference on Computer Vision and Pattern Recognition (CVPR) workshops

\bibitem[{Simon and Rodner(2015)}]{simon2015neural}
Simon M, Rodner E (2015) Neural activation constellations: Unsupervised part
  model discovery with convolutional networks. In: Proceedings of the IEEE/CVF
  International Conference on Computer Vision (ICCV)

\bibitem[{Simon et~al.(2014)Simon, Rodner, and Denzler}]{simon2014part}
Simon M, Rodner E, Denzler J (2014) Part detector discovery in deep
  convolutional neural networks. In: Asian Conference on Computer Vision

\bibitem[{Simonyan and Zisserman(2015)}]{simonyan2014very}
Simonyan K, Zisserman A (2015) Very deep convolutional networks for large-scale
  image recognition. ICLR

\bibitem[{Spivak(2012)}]{spivak2012outside}
Spivak GC (2012) Outside in the teaching machine. Routledge

\bibitem[{Sun et~al.(2018)Sun, Yuan, Zhou, and Ding}]{sun2018multi}
Sun M, Yuan Y, Zhou F, Ding E (2018) Multi-attention multi-class constraint for
  fine-grained image recognition. In: Proceedings of the European Conference on
  Computer Vision (ECCV), pp 805--821

\bibitem[{Szegedy et~al.(2016)Szegedy, Vanhoucke, Ioffe, Shlens, and
  Wojna}]{szegedy2016rethinking}
Szegedy C, Vanhoucke V, Ioffe S, Shlens J, Wojna Z (2016) Rethinking the
  inception architecture for computer vision. In: Conference on computer vision
  and pattern recognition, pp 2818--2826

\bibitem[{Tan and Le(2019)}]{tan2019efficientnet}
Tan M, Le Q (2019) Efficientnet: Rethinking model scaling for convolutional
  neural networks. In: International Conference on Machine Learning, PMLR, pp
  6105--6114

\bibitem[{Van~Horn et~al.(2015)Van~Horn, Branson, Farrell, Haber, Barry,
  Ipeirotis, Perona, and Belongie}]{van2015NaBirdDataset}
Van~Horn G, Branson S, Farrell R, Haber S, Barry J, Ipeirotis P, Perona P,
  Belongie S (2015) Building a bird recognition app and large scale dataset
  with citizen scientists: The fine print in fine-grained dataset collection.
  In: IEEE/CVF Conference on Computer Vision and Pattern Recognition (CVPR)

\bibitem[{Wah et~al.(2011)Wah, Branson, Welinder, Perona, and
  Belongie}]{WahCUBDataset}
Wah C, Branson S, Welinder P, Perona P, Belongie S (2011) The caltech-ucsd
  birds-200-2011 dataset. California Institute of Technology, Tech rep

\bibitem[{Wang et~al.(2015)Wang, Shen, Shao, Zhang, Xue, and
  Zhang}]{wang2015multiple}
Wang D, Shen Z, Shao J, Zhang W, Xue X, Zhang Z (2015) Multiple granularity
  descriptors for fine-grained categorization. In: Proceedings of the IEEE/CVF
  International Conference on Computer Vision (ICCV)

\bibitem[{Wegner et~al.(2016)Wegner, Branson, Hall, Schindler, and
  Perona}]{wegner2016cataloging}
Wegner JD, Branson S, Hall D, Schindler K, Perona P (2016) Cataloging public
  objects using aerial and street-level images-urban trees. In: IEEE/CVF
  Conference on Computer Vision and Pattern Recognition (CVPR)

\bibitem[{Xiao et~al.(2015)Xiao, Xu, Yang, Zhang, Peng, and
  Zhang}]{xiao2015application}
Xiao T, Xu Y, Yang K, Zhang J, Peng Y, Zhang Z (2015) The application of
  two-level attention models in deep convolutional neural network for
  fine-grained image classification. In: IEEE/CVF Conference on Computer Vision
  and Pattern Recognition (CVPR)

\bibitem[{Xie et~al.(2013)Xie, Tian, Hong, Yan, and
  Zhang}]{xie2013hierarchical}
Xie L, Tian Q, Hong R, Yan S, Zhang B (2013) Hierarchical part matching for
  fine-grained visual categorization. In: Proceedings of the IEEE/CVF
  International Conference on Computer Vision (ICCV)

\bibitem[{Xie and Tu(2015)}]{xie2015holistically}
Xie S, Tu Z (2015) Holistically-nested edge detection. In: Proceedings of the
  IEEE/CVF International Conference on Computer Vision (ICCV)

\bibitem[{Yang et~al.(2012)Yang, Bo, Wang, and Shapiro}]{yang2012unsupervised}
Yang S, Bo L, Wang J, Shapiro LG (2012) Unsupervised template learning for
  fine-grained object recognition. In: Advances in Neural Information
  Processing Systems (NIPS)

\bibitem[{Yang et~al.(2018)Yang, Luo, Wang, Hu, Gao, and
  Wang}]{yang2018learning}
Yang Z, Luo T, Wang D, Hu Z, Gao J, Wang L (2018) Learning to navigate for
  fine-grained classification. In: Proceedings of the European Conference on
  Computer Vision (ECCV)

\bibitem[{Zeiler and Fergus(2014)}]{zeiler2014visualizing}
Zeiler MD, Fergus R (2014) Visualizing and understanding convolutional
  networks. In: Proceedings of the European Conference on Computer Vision
  (ECCV)

\bibitem[{Zhang et~al.(2020)Zhang, Wu, Zhang, Zhu, Zhang, Lin, Sun, He, Muller,
  Manmatha, Li, and Smola}]{zhang2020resnest}
Zhang H, Wu C, Zhang Z, Zhu Y, Zhang Z, Lin H, Sun Y, He T, Muller J, Manmatha
  R, Li M, Smola A (2020) Resnest: Split-attention networks. arXiv preprint
  arXiv:200408955

\bibitem[{Zhang et~al.(2013)Zhang, Farrell, Iandola, and
  Darrell}]{zhang2013deformable}
Zhang N, Farrell R, Iandola F, Darrell T (2013) Deformable part descriptors for
  fine-grained recognition and attribute prediction. In: Proceedings of the
  IEEE/CVF International Conference on Computer Vision (ICCV)

\bibitem[{Zhang et~al.(2014)Zhang, Donahue, Girshick, and
  Darrell}]{zhang2014part}
Zhang N, Donahue J, Girshick R, Darrell T (2014) Part-based r-cnns for
  fine-grained category detection. In: Proceedings of the European Conference
  on Computer Vision (ECCV)

\bibitem[{Zhang et~al.(2016{\natexlab{a}})Zhang, Xiong, Zhou, Lin, and
  Tian}]{zhang2016picking}
Zhang X, Xiong H, Zhou W, Lin W, Tian Q (2016{\natexlab{a}}) Picking deep
  filter responses for fine-grained image recognition. In: IEEE/CVF Conference
  on Computer Vision and Pattern Recognition (CVPR)

\bibitem[{Zhang et~al.(2016{\natexlab{b}})Zhang, Zhou, Lin, and
  Zhang}]{zhang2016embedding}
Zhang X, Zhou F, Lin Y, Zhang S (2016{\natexlab{b}}) Embedding label structures
  for fine-grained feature representation. In: IEEE/CVF Conference on Computer
  Vision and Pattern Recognition (CVPR)

\bibitem[{Zhang et~al.(2016{\natexlab{c}})Zhang, Wei, Wu, Cai, Lu, Nguyen, and
  Do}]{zhang2016weakly}
Zhang Y, Wei XS, Wu J, Cai J, Lu J, Nguyen VA, Do MN (2016{\natexlab{c}})
  Weakly supervised fine-grained categorization with part-based image
  representation. IEEE Transactions on Image Processing

\bibitem[{Zheng et~al.(2017)Zheng, Fu, Mei, and Luo}]{zheng2017learning}
Zheng H, Fu J, Mei T, Luo J (2017) Learning multi-attention convolutional
  neural network for fine-grained image recognition. In: Proceedings of the
  IEEE/CVF International Conference on Computer Vision (ICCV)

\bibitem[{Zheng et~al.(2019)Zheng, Fu, Zha, and Luo}]{zheng2019TASN}
Zheng H, Fu J, Zha ZJ, Luo J (2019) Looking for the devil in the details:
  Learning trilinear attention sampling network for fine-grained image
  recognition. In: IEEE/CVF Conference on Computer Vision and Pattern
  Recognition (CVPR)

\bibitem[{Zoph et~al.(2018)Zoph, Vasudevan, Shlens, and Le}]{zoph2018learning}
Zoph B, Vasudevan V, Shlens J, Le QV (2018) Learning transferable architectures
  for scalable image recognition. In: IEEE/CVF Conference on Computer Vision
  and Pattern Recognition (CVPR)

\end{thebibliography}

\end{document}